\documentclass[letterpaper]{article} 
\usepackage{aaai24}  
\usepackage{times}  
\usepackage{helvet}  
\usepackage{courier}  
\usepackage[hyphens]{url}  
\usepackage{graphicx} 
\usepackage{natbib}  
\usepackage{caption} 
\usepackage{algorithm}
\usepackage{algorithmic}
\usepackage{multirow}
\usepackage{array}
\usepackage{booktabs}
\usepackage{adjustbox}
\usepackage{amssymb}
\usepackage{graphicx}
\usepackage{subcaption}
\usepackage{tikz}

\usepackage{xcolor,colortbl}
\definecolor{Gray}{gray}{0.85}

\usepackage{amsmath}
\usepackage{newfloat}
\usepackage{listings}
\usepackage{bm}
\DeclareCaptionStyle{ruled}{labelfont=normalfont,labelsep=colon,strut=off} 
\floatstyle{ruled}
\newfloat{listing}{tb}{lst}{}
\floatname{listing}{Listing}
\title{Root Cause Analysis In Microservice Using Neural Granger Causal Discovery}
\author{
    Cheng-Ming Lin, Ching Chang, Wei-Yao Wang, Kuang-Da Wang, Wen-Chih Peng
}
\affiliations{
    National Yang Ming Chiao Tung University, Hsinchu, Taiwan \\
    zmlin1998.cs10@nycu.edu.tw, blacksnail789521@gmail.com, sf1638.cs05@nctu.edu.tw, gdwang.cs10@nycu.edu.tw, wcpeng@cs.nycu.edu.tw
}

\begin{document}

\maketitle

\begin{abstract}
In recent years, microservices have gained widespread adoption in IT operations due to their scalability, maintenance, and flexibility. 
However, it becomes challenging for site reliability engineers (SREs) to pinpoint the root cause due to the complex relationships in microservices when facing system malfunctions.
Previous research employed structured learning methods (e.g., PC-algorithm) to establish causal relationships and derive root causes from causal graphs.
Nevertheless, they ignored the temporal order of time series data and failed to leverage the rich information inherent in the temporal relationships.
For instance, in cases where there is a sudden spike in CPU utilization, it can lead to an increase in latency for other microservices.
However, in this scenario, the anomaly in CPU utilization occurs before the latency increase, rather than simultaneously. 
As a result, the PC-algorithm fails to capture such characteristics.
To address these challenges, we propose RUN, a novel approach for root cause analysis using neural Granger causal discovery with contrastive learning.
RUN enhances the backbone encoder by integrating contextual information from time series, and leverages a time series forecasting model to conduct neural Granger causal discovery.
In addition, RUN incorporates Pagerank with a personalization vector to efficiently recommend the top-k root causes.
Extensive experiments conducted on the synthetic and real-world microservice-based datasets demonstrate that RUN noticeably outperforms the state-of-the-art root cause analysis methods.
Moreover, we provide an analysis scenario for the sock-shop case to showcase the practicality and efficacy of RUN in microservice-based applications.
Our code is publicly available at https://github.com/zmlin1998/RUN.
\end{abstract}

\section{Introduction}

Root cause analysis plays a crucial role in numerous domains such as cloud system operations \cite{DBLP:conf/dais/Soldani0B22}, manufacturing processes \cite{DBLP:journals/jim/OliveiraMB23}, or telecommunications networks \cite{DBLP:journals/wias/ChenYLWLTHW22}.
For instance, during the manufacturing process of wafers, if the thickness of a wafer is found to deviate from the standard, root cause analysis can be employed to identify the specific manufacturing step that caused the abnormal thickness.
By applying root cause analysis to sensor data with complex relationships, these situations can effectively pinpoint the underlying causes when a system malfunction occurs.

In recent years, as the companies experience continued growth, the system operations have scaled up and grown increasingly complex.
Therefore, these organizations have opted to migrate from the so-called monolithic architecture to microservice architecture \cite{DBLP:conf/icse/LiuH0LZGLOW21}. 
Microservice architecture offers benefits such as better scalability, easier maintenance, and greater flexibility. 
Each service can be independently scaled based on demand, enabling efficient resource utilization.
Developers can focus on individual services, making it simpler to debug, test, and deploy changes.
Despite the numerous benefits of microservices, when an anomaly arises in one service, the interdependencies among services can create a domino effect, resulting in subsequent issues and ultimately leading to system failure \cite{DBLP:conf/icse/Zhang0SZFWL022}. 
In such scenarios, in-depth analysis becomes imperative to identify the culprit of the anomaly and mitigate the problem effectively.

\begin{figure}[tbp]
  \centering
  \includegraphics[width=1\linewidth]{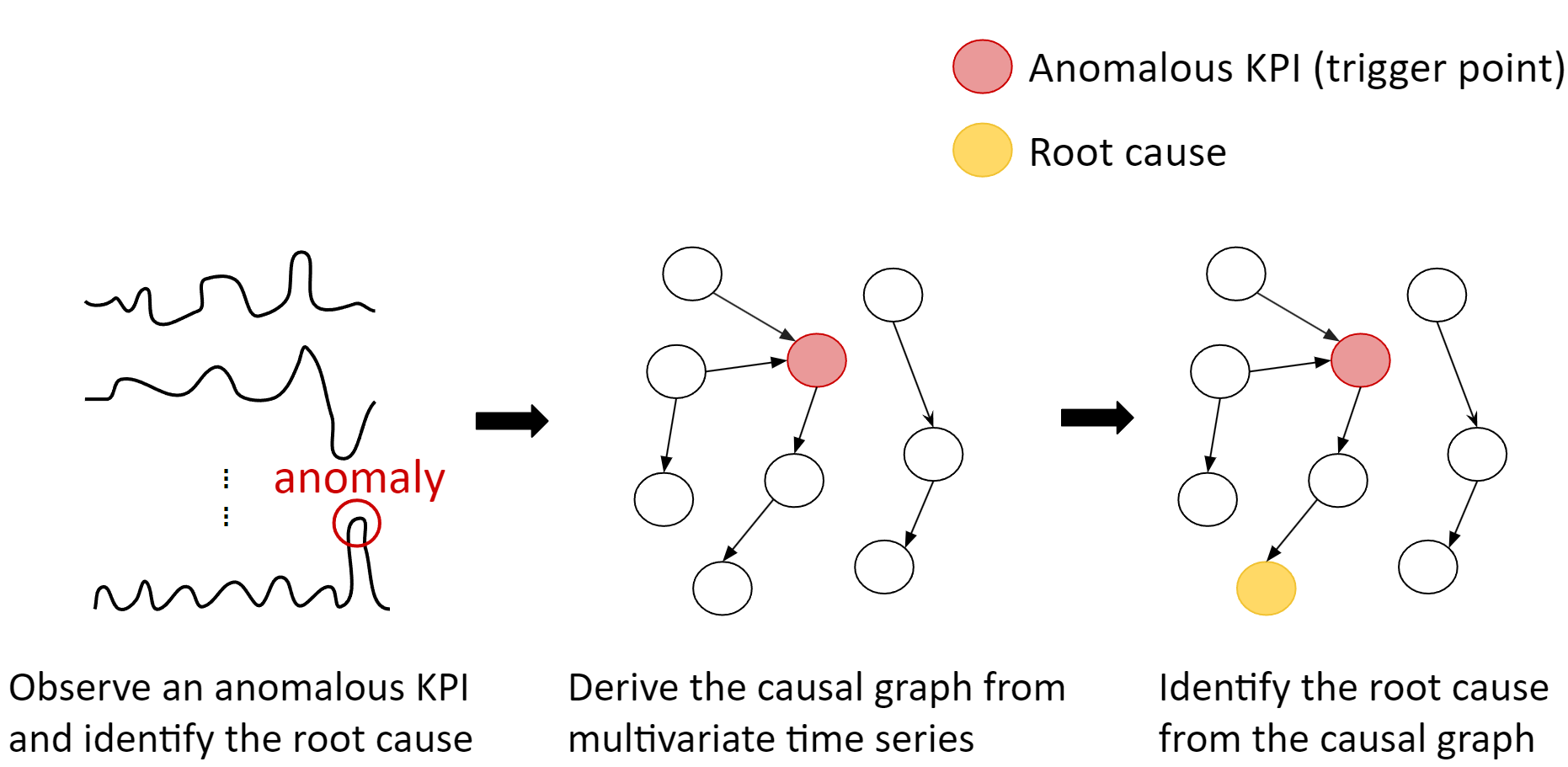}
  \caption{An example of causal structure discovery-based techniques for RCA.
  }
  \label{fig:RCA}
\end{figure}

However, in microservice monitoring systems, only the operational values of the system are recorded, without documenting the relationships between them.
Hence, researchers have recently employed causal structure discovery-based techniques for Root Cause Analysis (RCA) in cloud applications \cite{DBLP:journals/corr/abs-2302-01987}, aiming to identify the underlying causes of anomalies.
Figure \ref{fig:RCA} illustrates a flow of RCA using the causal structure discovery-based approach.
When the anomaly detection alarm is triggered by an anomalous Key Performance Indicator (KPI), engineers initially designate that particular KPI as the \textbf{trigger point}. 
Subsequently, they aim to identify the underlying root cause of this trigger point. 
To achieve this, they construct a causal graph that establishes relationships between different KPIs, allowing them to precisely pinpoint the culprit of the anomaly using the insights provided by the causal graph.


On the other hand, the Granger causality \cite{granger1969investigating} analysis is another widely recognized approach used to assess whether a set of time series $x$ is a causal factor for another set of time series $y$.
Granger causality has gained attention and is widely acknowledged for its advantages of interpretability and compatibility with the emergence of deep learning \cite{DBLP:journals/pami/TankCFSF22}.
Previous works \cite{DBLP:journals/make/NautaBS19} utilized time series forecasting models and interpreted model parameters as the relationships between variables to establish causal relationships.
However, prior approaches in neural Granger causal discovery have not effectively leveraged the contextual information inherent in temporal data.

The previous approach \cite{DBLP:conf/aaai/YueWDYHTX22} generated positive and negative samples at instance-wise and temporal dimensions for generating fine-grained representations for any granularity.
However, we notice that real-world time series often exhibit multiple-periodicity.
For instance,  the temperature tends to reach its peak at noon every day, there is an influx of people in the city during weekends, and electricity consumption significantly increases during the summer every year.
Therefore, this approach might inadvertently treat timestamps with the same periodicity as negative pairs, even though they actually hold similar significance.

To tackle these challenges, we propose a novel framework \textbf{RUN}, which employs a self-supervised neural Granger causal discovery approach for conducting root cause analysis. 
To capture contextual information in time series, we employ a self-supervised learning scheme that ensures timestamps with diverse contexts, but the same timestamps are learned to exhibit proximity, where we only treat the identical timestamps as positive pairs for contrastive learning.
This way aims to prevent the inclusion of erroneous information stemming from negative pairs.
After enhancing the backbone encoder to acquire improved representations, we utilize neural Granger causal discovery to explore the causal relationships among variables.
Subsequently, we proceed to identify the root causes of the trigger point within the acquired causal graph using Pagerank with a personalized vector.
Comprehensive experiments conducted on both synthetic and real-world microservice-based datasets conclusively demonstrate that our proposed framework outperforms state-of-the-art root cause analysis methods.

Our main contributions are summarized as follows:
\begin{itemize}
    \item We propose a self-supervised neural Granger causal discovery-based root cause analysis framework \textbf{RUN}, which captures contextual information in temporal data, and then leverages the time series forecasting model to construct a causal graph between multivariate time series. \textbf{RUN} employs Pagerank on the derived causal graph to identify the root cause of the trigger point.
    \item We introduce an innovative self-supervised learning method for time series data that exclusively treats identical timestamps with distinct contextual information as positive pairs. 
    This approach mitigates the problem of misidentifying timestamps with similar periodicity as negative pairs, thereby preventing the separation of representations for timestamps sharing the same periodicity.
    \item \textbf{RUN} outperforms existing state-of-the-art methods on synthetic datasets and data generated from a real-world application. We further conduct an ablation study to validate the effectiveness of our contrastive learning method.
\end{itemize}

\section{Related Work}

\begin{figure}
  \includegraphics[width=1.0\linewidth]{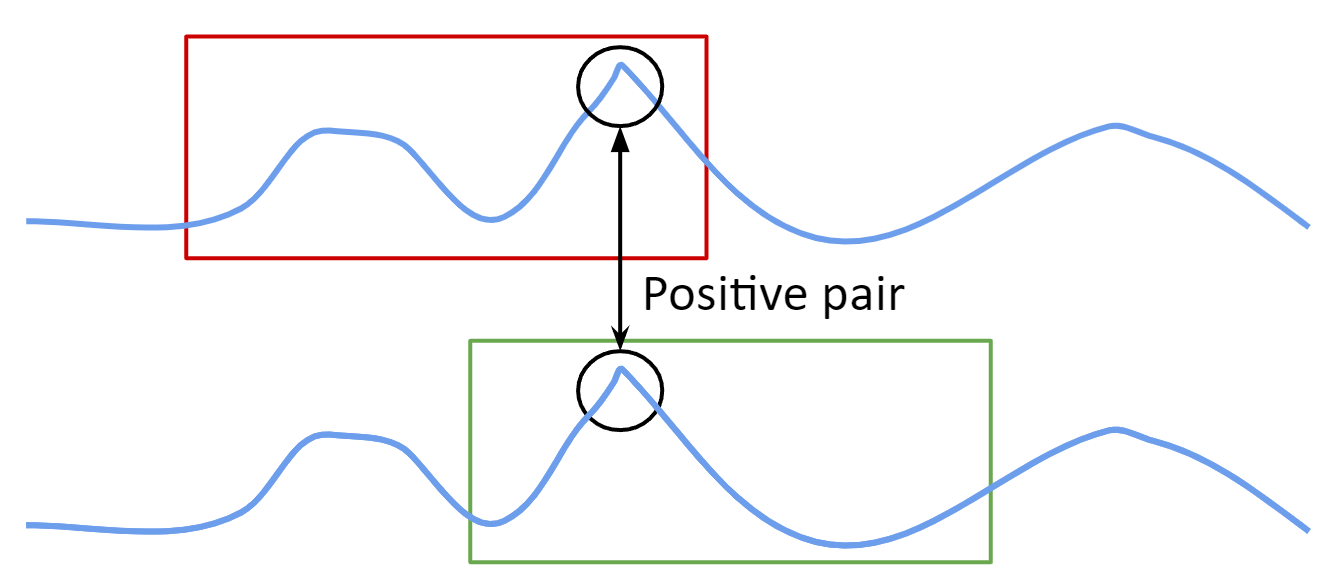}
  \caption{Illustrations of contextual information from the same timestamps but with different contexts. The red window and green window respectively represent two distinct types of contextual information. The same timestamp with different contexts should be close.}
  \label{fig:positive pair}
\end{figure}

\begin{figure}[tbp]
  \centering
  \includegraphics[width=1.0\linewidth]{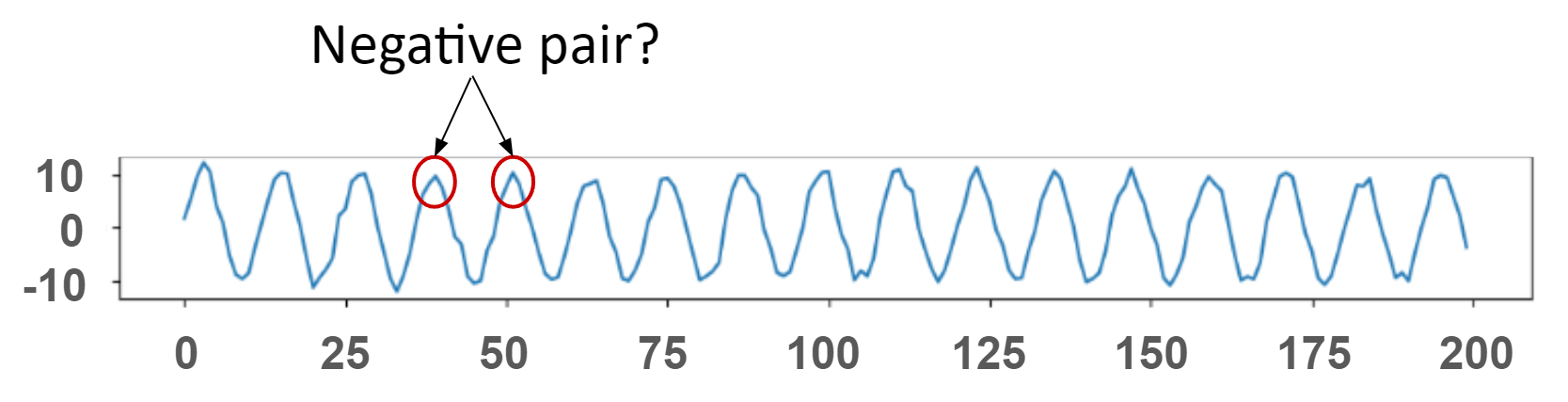}
  \caption{An illustrated issue of negative pair selection.}
  \label{fig:negative}
\end{figure}

\begin{figure*}[htp]
  \centering
  \includegraphics[width=1.0\linewidth]{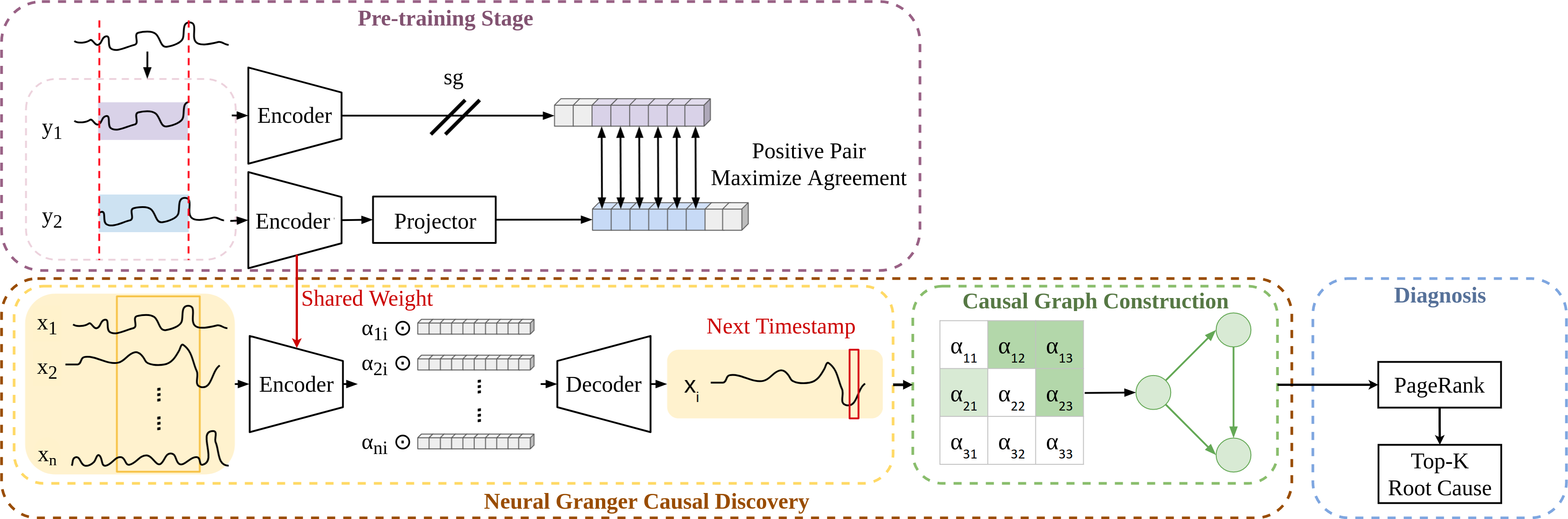}
  \caption{Overview of our proposed framework, RUN, consisting of three stages: 1) Maximizing the positive pair to capture the contextual information; 2) Neural Granger causal discovery to derive the causal graph from multivariate time series; and 3) The diagnosis stage infers the root cause from the obtained causal graph.}
  \label{fig:framework}
\end{figure*}
\subsection{Root Cause Analysis in Microservices}
Within large organizations, the adoption of microservices has become the prevailing architecture due to its advantages in terms of enhanced scalability, simplified maintenance, and increased flexibility.
When the anomaly detection alarm is activated by an anomalous KPI, engineers often have to invest a significant amount of time to pinpoint the underlying cause behind this anomaly.
Consequently, they engage in the development of automated root cause analysis systems to mitigate the time spent on the investigation.
$\epsilon$-diagnosis \cite{DBLP:conf/www/ShanCLZXHLD19} considers the time series of KPIs and computes the similarity between them during normal periods and abnormal periods, which serves as an anomaly indicator.
AutoMAP \cite{DBLP:conf/www/MaXWCZW20} utilizes the PC-algorithm to construct a causality graph based on the KPIs.
It further conducts a random walk to determine the candidate's root causes.
RCD \cite{DBLP:conf/nips/IkramCMSBK22} employs the hierarchical and localized learning algorithm.
It deems the failures as an intervention, and adopts a divide-and-conquer scheme to divide the variables into multiple subsets for conducting $\Psi$-PC to find interventional targets from each subset.

The candidate root causes from all subsets are then combined, and the same process is iteratively applied until further splitting of candidate root causes is not possible.
Nonetheless, these methods rely on similarity or the PC-algorithm, both of which overlook the essential role of temporal dependency in root cause analysis.
For instance, when a sudden spike in CPU utilization occurs, it can lead to increased latency in other microservices, effectively encompassing the temporal dimension of the information.
Therefore, we aim to develop a neural Granger causal discovery method that can leverage the temporal dependency to identify the root causes more accurately.

\subsection{Neural Granger Causal Discovery}
As deep Neural Networks (NNs) continue to advance rapidly, researchers have started using Recurrent Neural Networks or other Temporal Convolutional Networks to infer nonlinear Granger causality.
MSNGC \cite{fan2023interpretable} extracts diverse causal information from the data, considering various delay ranges, and effectively integrates them using learned attention weights.
\citet{DBLP:journals/pami/TankCFSF22} applies structured component-wise multilayer perceptrons (MLPs) combined with sparsity-inducing penalties on the weights.
CUTS \cite{DBLP:conf/iclr/ChengYXLSHD23} utilizes EM-Style joint causal graph learning and missing data imputation for irregular temporal data.
Nevertheless, previous works have not effectively leveraged contextual information in time series data.
As a result, we present an innovative approach for time series contrastive learning that enables to the capture of contextual information from identical timestamps but with different contexts, as depicted in Figure \ref{fig:positive pair}.

\subsection{Contrastive Learning}
Contrastive learning has been widely used in various domains, such as natural language processing (NLP) \cite{DBLP:conf/emnlp/GaoYC21}, computer vision (CV) \cite{DBLP:conf/cvpr/ChenH21}, and time series \cite{DBLP:conf/aaai/YueWDYHTX22}.
TS2Vec \cite{DBLP:conf/aaai/YueWDYHTX22} samples positive pairs which are the same timestamp with different contexts, and negative pairs which are different timestamps.
However, we observe that real-world time series frequently demonstrate multiple-periodicity.
With the advancement of contrastive learning, \cite{DBLP:conf/nips/GrillSATRBDPGAP20, DBLP:conf/cvpr/ChenH21} argue that it can perform well even without negative pairs because they aim to prevent the inclusion of incorrect negative pairs.
Hence, we propose an innovative approach that exclusively leverages identical timestamps as positive pairs for self-supervised learning. 
This scheme is designed to prevent the incorporation of information from negative pairs.
In Figure \ref{fig:negative}, we can observe that the time series data exhibits periodic patterns.
Therefore, if we were to consider the two circled red points on the graph as negative pairs, it would lead to incorporating incorrect information from them.

\section{Problem Formulation}
The monitoring system will monitor the operational status of the microservice system through KPIs, and in typical scenarios, we pinpoint the root cause when an anomalous KPI triggers the anomaly detection system. 
We term the anomalous KPI as a \textbf{trigger point} and subsequently identify the underlying root causes behind the trigger point.
These KPIs belong to multivariate time series data.
In the microservice architecture, identifying the underlying causes for the anomaly can be formalized as follows. 
Multivariate time series $\textbf{X}$ consists of $N$ features, $\textbf{X} = [X_1, X_2, ..., X_N]$.
We collect the corresponding time series with a specified time period $T$, each time series can be denoted as $X_i = [x^1_i, x^2_i, \ldots, x^T_i]$, $\textbf{X} \in \mathbb{R}^{N \times T}$. 
We utilize the neural Granger causal discovery method to derive the causal graph $\Hat{\textbf{G}} = \{V, E\}$, where nodes $V$ denotes each feature $X_i$ and edges $E$ denotes that feature $X_i$ Granger causes feature $X_j$.
Our goal is to identify the root cause $X_{culprit}$ which leads to the trigger point within the causal graph $\Hat{\textbf{G}}$.

\section{Methodology}

Our proposed framework is shown in Figure \ref{fig:framework}, which consists of three stages: the pre-training stage, the neural Granger causal discovery stage, and the diagnosis stage.
In the pre-training stage, we enhance the backbone encoder to generate informative time series representations that incorporate contextual information.
This enhancement is achieved by maximizing the agreement among instances with the same timestamp but with different contexts.
In the neural Granger causal discovery stage, we utilize the time series forecasting model to exploit the causal graph and prune the spurious edge to get the final Directed Acyclic Graphs (DAGs).
In the diagnosis stage, we apply a random walk algorithm called Pagerank with a personalization vector which recommends the most likely root causes based on the causal graph.

\subsection{Pre-training Stage}
We utilize DLinear \cite{DBLP:conf/aaai/ZengCZ023} as our backbone encoder for time series forecasting.
In order to enhance our backbone encoder, we design contrastive learning without relying on negative pairs.
We consider that negative pairs are not suitable for time series because they exhibit periodic patterns as depicted in Figure \ref{fig:negative}.
When selecting different timestamps as negative pairs, it is possible that they belong to the same periodicity.
If we choose the incorrect negative pairs, it may lead to separating 
 embeddings that should be similar.
Therefore, we learn the context information by considering instances with the same timestamp but with different contexts as positive pairs.

First, we utilize random cropping following \cite{DBLP:conf/aaai/YueWDYHTX22} by randomly sampling two overlapping time segments $y_1$ and $y_2$, 
\begin{equation}
    y_1 = [a_1, b_1], y_2 = [a_2, b_2], 0 < a_1 \leq a_2 \leq b_1 \leq b_2 \leq T.
\end{equation}
The two views $y_1$ and $y_2$ are processed by our backbone encoder $f$ and a projector $g$.
To learn the contextual information without negative pairs, we utilize the same schema of SimSiam \cite{DBLP:conf/cvpr/ChenH21}.
We denote the two representations as $z_1 = g(f(y_1))$ and $p_2 = f(y_2)$, and maximize the similarity between both sides.
Here, we define our loss function as follows:
\begin{equation}
\small
    L_{con} = - \frac{1}{2} (cos(z_1, stopgrad(p_2)) + cos(stopgrad(p_1), z_2)). 
\end{equation}

\subsection{Neural Granger Causal Discovery Stage}
RUN represents the causal graph as a Causal Attention Matrix $\textbf{G} = \{0 \leq \alpha_{ij} \leq 1, \ \forall i,j \in [1,2, ..., N] \}$, where the element $\alpha_{ij}$ demonstrates the attention score from $X_i$ to $X_j$.
Each element $\alpha_{ij}$ serves as a trainable parameter, contributing to identifying causal relationships among time series.
We define the final causal graph $\bm{\widetilde{G}} = \{g_{ij} \in [0, 1], \forall i,j \in [0, 1]\}$. 
If the attention score $\alpha_{ij}$ is beyond a certain threshold, $g_{ij}$ will be 1, indicating that time series $X_i$ Granger causes $X_j$.
The proposed neural Granger causal discovery comprises two components: time series forecasting and causal graph discovery.

\subsubsection{Time Series Forecasting}
\begin{figure}[tbp]
  \includegraphics[width=1\linewidth]
  {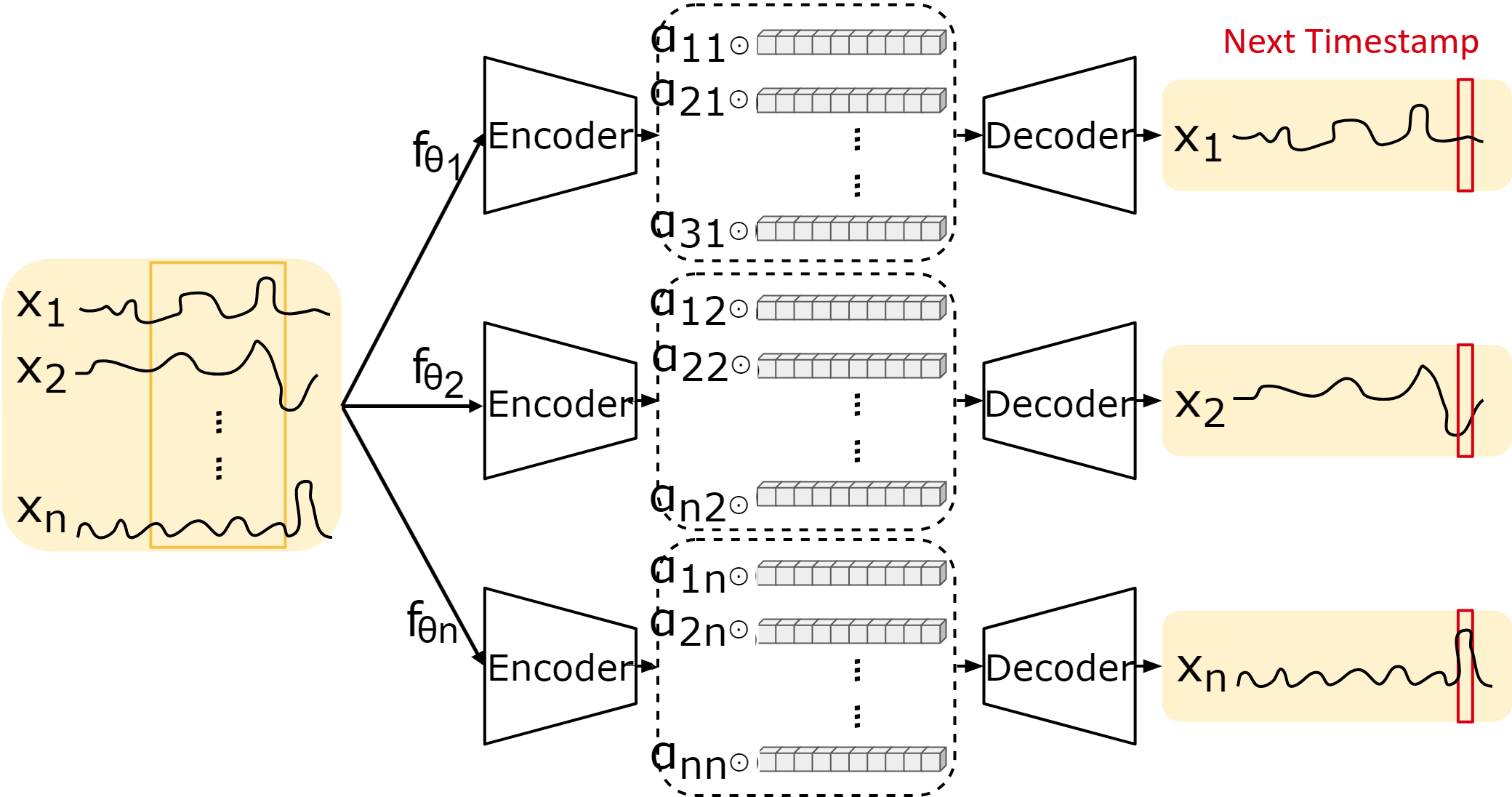}
  \caption{Overview of time series forecasting. There are $N$ independent neural networks for each time series $i$ to predict their causal relationships.}
  \label{fig:timeseries}
\end{figure}

In time series forecasting, we propose $N$ independent neural network $f_{\theta _i}$ for time series $X_i$ to fit its data generation function \cite{DBLP:conf/iclr/ChengYXLSHD23}.
We define the input for our model by applying a sliding window with a size of $w$ across all the historical time series $\textbf{X}$ and multiplying with corresponding attention score $\alpha_{ni}$ and the forecast $\hat{x}_i$ at timestamp $t$ is the output of the neural network $f_{\theta _i}$
\begin{equation}
\begin{aligned}
    \hat{x}^t_i = f_{\theta _i}(\textbf{X} \odot G) 
                = f_{\theta _i}(x^{t-w:t-1}_1 \odot \alpha_{1i}
                , ...,x^{t-w:t-1}_N \odot\alpha_{Ni}), \\
                i = 1, 2, ...,N.
\end{aligned}
\end{equation}
An overview of time series forecasting is illustrated in \ref{fig:timeseries}.
Each independent neural network has the same architecture but a different target time series.
For a neural network $f_{\theta _i}$, the goal is to forecast corresponding time series $X_i$ by minimizing the following loss function
\begin{equation}
    L_{MSE} = - \frac{1}{N-w} \sum_{t=w+1}^{T} (x^t_i - \hat{x}^t_i)^2
\end{equation}

\subsubsection{Causal Graph Discovery}
After training the networks, the learnable Causal Attention Matrix $G$ can be used to interpret the causal relationships among time series.
When $\alpha_{ij}$ exceeds a certain threshold $H$, we can infer that time series $X_i$ Granger causes $X_j$ and add an edge from $X_i$ to $X_j$ in the causal graph $\bm{\widetilde{G}}$, that is
\begin{equation}
    \bm{\widetilde{G}}_{ij} = \begin{cases}
    1, & \text{if} \ \alpha_{ij} > H \\
    0, & \text{else} \ 
    \end{cases}
\end{equation}
Based on the definition of a causal graph, cycles are not allowed in the graph. 
Therefore, we need to incorporate a pruning strategy to remove spurious edges.
We calculate the similarity of each edge, which involves computing the Pearson correlation between two connected nodes.
Subsequently, we eliminate the edge with the lowest similarity iteratively until $\bm{\widetilde{G}}$ transforms into the final causal graph $\Hat{\textbf{G}}$.

\subsection{Diagnosis Stage}
In the diagnosis stage, we follow GrootRank \cite{DBLP:conf/kbse/WangWJHWKX21} to apply PageRank with node weight personalization to calculate the root cause ranking.
\citet{DBLP:conf/kbse/WangWJHWKX21} observes that dangling nodes are more likely to be the root cause.
Hence, we customize the personalization vector as $P_d$ and $P_n$, where $P_d$ is the personalization score for dangling nodes and $P_n$ is for the remaining nodes.
In the case of tied rankings, we calculate the access distance from the trigger point to resolve the tie.
We calculate the access distance (AD) differently from GrootRank
\begin{equation}
    \bm{AD} = \begin{cases}
    D, & \text{distance from trigger point to the variable}\\
    0, & \text{if any “access" is not reachable} \ 
    \end{cases}
\end{equation}
In GrootRank, the authors set the distance of unreachable nodes as infinity and consider that shorter access distance will more likely be the root cause.
However, this may contradict the observed phenomena which suggest that dangling nodes are more likely to be the root cause.
Hence, we set the distance of the unreachable node as $0$ and consider the bigger access distance will more likely be the root cause.
Finally, RUN outputs the top-k root causes based on the root cause ranking.

\section{Experiments}

\subsection{Experiment Settings}
\subsubsection{Dataset}
As no publicly available real-world dataset for root cause analysis is accessible due to data confidentiality, we test on a synthetic dataset and a test bed utilizing an actual microservice-based application.
\begin{itemize}
    \item Synthetic data: To generate synthetic data, we follow RCD \citet{DBLP:conf/nips/IkramCMSBK22} to randomly generate DAG and generate conditional probability tables (CPT) for the normal period dataset.
    Then, we inject failures by randomly choosing one node as the root cause and regenerating CPT for it.
    Hence, we can get the anomalous period dataset and combine them into one case.
    Finally, we also utilize an anomaly detection system to identify the trigger point.
    The data are generated with node counts of 10, 20, 30, 40, and 50, spanning over 2,000 timestamps.
    \item Sock-shop \cite{githubrepo}: The framework of sock-shop encompasses a total of 13 microservices, each developed using distinct technologies. 
    These microservices are deployed on individual virtual machines or containers. 
    Communication among them occurs through HTTP-based API requests. Additionally, these microservices offer a substantial volume of statistical data in the form of various metrics such as CPU and memory utilization.
    Within the sock-shop dataset, ten distinct categories of root causes are present, each comprising five instances. 
\end{itemize}

\subsubsection{Baselines}
We compare our performance with the following baseline approaches:
\begin{itemize}
    \item $\epsilon$-Diagnosis \cite{DBLP:conf/www/ShanCLZXHLD19}: $\epsilon$-Diagnosis analyzes the time series of KPIs and calculates the similarity between them during both normal and abnormal periods, which serves as an indicator for anomalies. 
    If the similarity of a particular KPI between normal and anomalous time periods falls below a certain confidence threshold, indicating significant changes in that KPI before and after the anomaly occurrence, it can be considered a candidate root cause.
    \item AutoMAP \cite{DBLP:conf/www/MaXWCZW20}: AutoMAP creates a weighted causal graph utilizing an adapted PC algorithm. 
    In this graph, the weight of an edge between two microservices reflects the degree of dependence of a performance metric. 
    The process concludes by navigating nodes through a random walk algorithm and identifying the root cause via the correlation metric from the weighted causal graph.
    \item RCD \cite{DBLP:conf/nips/IkramCMSBK22}: 
    RCD treats failures as interventions on the nodes representing root causes, and utilizes a localized hierarchical learning algorithm to identify these root causes of failure. 
    It also employs a divide-and-conquer scheme to decrease the time required for inferring the root cause from the entire graph.
    \item $\Psi$-PC: 
    $\Psi$-PC is a specific instance of RCD.
    $\Psi$-PC will acquire the entire causal graph, which might not be essential for root cause identification. Additionally, localizing learning would demand extra time.
    \item CausalRCA \cite{DBLP:journals/jss/XinCZ23}: CausalRCA uses a gradient-based causal structure learning method to generate weighted causal graphs and a root cause inference method to localize root cause metrics.
\end{itemize}

\begin{table}
    \begin{center}
\begin{tabular}{c|c|c}
    \toprule
    Dataset & \# of Time Series & \# of Timestamps \\
    \midrule
    \multirow{5}{*}{Synthetic data} 
    & 10 & \multirow{5}{*}{2000}\\
    & 20 \\
    & 30 \\
    & 40 \\
    & 50 \\
    \midrule
    Sock-shop & 38 & 600\\
    \bottomrule
\end{tabular}
\end{center} 
    \caption{Statistics of the synthetic and sock-shop datasets.}
    \label{tab:dataset}
\end{table}

\begin{table*}
    \begin{adjustbox}{center}
\resizebox{\textwidth}{!}{
\begin{tabular}{cl|cccccc|cccccc}
    \toprule
    \multicolumn{2}{c|}{\multirow{2}{*}{}}  & 
    \multicolumn{6}{c|}{CPU hog} & \multicolumn{6}{c}{Memory leak} \\
    \cline{3-14}
    & & $\epsilon$-Diag. & AMAP & $\Psi$-PC* & RCD* & CausalRCA & RUN & $\epsilon$-Diag. & AMAP & $\Psi$-PC* & RCD* & CausalRCA & RUN\\
    \midrule
    \multirow{6}{*}{HR@1} &  Carts  & \textbf{0.20} & \underline{0.05} & 0.02 & 0.00 & 0.00 & \textbf{0.20} & 0.00 & \textbf{0.14} & \underline{0.02} & \underline{0.02} & 0.00 & 0.00\\
    & Catalogue & 0.20 & \underline{0.22} & 0.02 & 0.08 & 0.00 &\textbf{0.40} & \textbf{0.40} & 0.00 & \underline{0.10} & 0.06 & 0.00 & 0.00\\
    & Orders & \underline{0.20} & 0.01 & 0.06 & 0.12 & \textbf{0.80} & 0.00 & 0.00 & \textbf{0.26} & 0.02 & 0.04 & 0.00 & \underline{0.20} \\
    & Payment & \textbf{0.60} & 0.07 & 0.06 & 0.04 & 0.00 & \underline{0.40} & \textbf{0.20} & 0.01 & \underline{0.02} & \underline{0.02} & 0.00 & 0.00\\
    & User & \underline{0.20} & \textbf{0.29} & 0.06 & 0.12 & \underline{0.20} & \underline{0.20} & \textbf{0.40} & 0.08 & \underline{0.20} & 0.14 & 0.00 & \textbf{0.40}\\
    \rowcolor{Gray}
    & Avg. & \underline{0.28} & 0.13 & 0.04 & 0.07 & 0.04 & \textbf{0.40} & \textbf{0.20} & 0.10 & 0.07 & 0.06 & 0.00 & \underline{0.12}\\
    \midrule
    \multirow{6}{*}{HR@3} &  Carts  & \textbf{0.40} & \underline{0.23} & 0.04 & 0.04 & 0.00 & \textbf{0.40} & 0.00 & \textbf{0.20} & 0.02 & \underline{0.06} & 0.00 & \textbf{0.20}\\
    & Catalogue & 0.20 & \underline{0.22} & 0.04 & 0.12 & 0.00 & \textbf{0.60} & \textbf{0.40} & 0.00 & 0.06 & \underline{0.12} & 0.00 & \textbf{0.40}\\
    & Orders & 0.20 & \underline{0.36} & 0.06 & 0.30 & 0.00 & \textbf{0.80} & 0.00 & \textbf{0.37} & 0.06 & 0.14 & 0.00 & \underline{0.20} \\
    & Payment & \textbf{0.60} & 0.07 & 0.06 & 0.12 & 0.00 & \underline{0.40} & \underline{0.20} & 0.05 & 0.02 & 0.16 & 0.00 & \textbf{0.40}\\
    & User & 0.20 & \textbf{0.53} & 0.04 & 0.12 & \underline{0.40} & 0.20 & \textbf{0.40} & 0.27 & 0.16 & \underline{0.28} & 0.20 & \textbf{0.40}\\
    \rowcolor{Gray}
    & Avg. & \underline{0.32} & 0.28 & 0.05 & 0.16 & 0.08 & \textbf{0.48} & \underline{0.20} & 0.14 & 0.06 & 0.15 & 0.04 & \textbf{0.32}\\
    \midrule
    \multirow{6}{*}{HR@5} &  Carts  & \textbf{0.40} & \underline{0.24} & 0.08 & 0.22 & 0.20 & \textbf{0.40} & 0.00 & \textbf{0.25} & 0.02 & 0.04 & \underline{0.20} & \underline{0.20}\\
    & Catalogue & 0.20 & 0.23 & 0.06 & 0.08 & \underline{0.40} & \textbf{0.80} & \underline{0.40} & 0.00 & 0.08 & 0.12 & 0.00 & \textbf{0.60}\\
    & Orders & 0.20 & \underline{0.51} & 0.06 & 0.36 & 0.00 & \textbf{0.80} & 0.00 & \textbf{0.49} & 0.08 & 0.08 & 0.00 & \underline{0.20} \\
    & Payment & \textbf{0.60} & 0.07 & 0.06 & 0.06 & 0.00 & \underline{0.40} & \underline{0.20} & 0.11 & 0.04 & 0.10 & \underline{0.20} & \textbf{0.40}\\
    & User & 0.20 & \textbf{0.55} & 0.06 & 0.34 & \underline{0.40} & 0.20 & \textbf{0.40} & \underline{0.30} & 0.20 & 0.28 & 0.20 & \textbf{0.40} \\
    \rowcolor{Gray}
    & Avg. & \underline{0.32} & \underline{0.32} & 0.06 & 0.21 & 0.20 & \textbf{0.52} & \underline{0.20} & 0.18 & 0.08 & 0.12 & 0.12 & \textbf{0.36}\\
    \bottomrule
\end{tabular}}
\end{adjustbox}

    \caption{HR@k on sock-shop data. Bold indicates the best performance and underline represents the second best performance. We note that $\Psi$-PC* and RCD* indicate different results compared with the original paper, where we directly reran the experiments from the official codes.}
    \label{tab:sockshop}
\end{table*}

\subsubsection{Evaluation Metrics}
We evaluate our solution by top-k hit ratio (HR@$k$) and mean reciprocal rank (MRR). 
\begin{itemize}
    \item HR@$k$ represents the probability of getting the correct root cause from the top-k outputs.
    \item MRR sums up the reciprocal of the rank of the root cause. If the root cause is not included in the output, its rank can be deemed as infinity, resulting in a score of zero.
\end{itemize}

\subsubsection{Implementation Details}
We implement our method on a machine with AMD EPYC 7302 16-Core CPU, NVIDIA RTX A5000 graphics cards.
In the time series forecasting stage, the window size $w$ is set to $32$.
We use the Adam \cite{DBLP:journals/corr/KingmaB14} optimizer and set the learning rate as $0.001$ and the batch size as $128$.
In the causal graph discovery stage, the threshold $H$ is set to $0.5$.
The training epochs of the pre-training and fine-tuning stages are set to 50.
In the diagnosis stage, we set the value of the personalization vector $P_d$ as 1 and $P_n$ as 0.5, similar to \cite{DBLP:conf/kbse/WangWJHWKX21}.
The k in $HR@k$ is set to 1, 3, and 5.

\begin{table}    
    \begin{adjustbox}{center}
\begin{tabular}{cl|cc|cc}
    \toprule
    \multicolumn{2}{c|}{\multirow{2}{*}{}}  & 
    \multicolumn{2}{c|}{CPU hog} & \multicolumn{2}{c}{Memory leak} \\
    \cline{3-6}
    & & $\epsilon$-Diag. & RUN & $\epsilon$-Diag. & RUN\\
    \midrule
    \multirow{6}{*}{MRR} &  Carts  & 0.273 & \textbf{0.299} & 0.036 & \textbf{0.158} \\
    & Catalogue & 0.203 & \textbf{0.537} & \textbf{0.412} & 0.207 \\
    & Orders & 0.212 & \textbf{0.825} & 0.017 & \textbf{0.267} \\
    & Payment & \textbf{0.613}& 0.494 & \textbf{0.437} & 0.164 \\
    & User & 0.221 & \textbf{0.264} & 0.231 & \textbf{0.443} \\
    \rowcolor{Gray} 
    & Avg. & 0.304 & \textbf{0.484} & 0.227 & \textbf{0.248} \\
    \bottomrule
\end{tabular}
\end{adjustbox}

    \caption{MRR on sock-shop data.}
    \label{tab:MRR sock shop}
\end{table}

\subsection{Overall Performance}
\subsubsection{Sock-shop}
Table \ref{tab:sockshop} demonstrates the HR@k of root cause analysis methods between RUN and baselines, which demonstrates that our proposed model achieves at least a 63\% improvement in CPU hog, and 80\% improvement in Memory leak compared with the best-performing baseline.
Considering baselines, $\epsilon$-diagnosis outperforms other baselines because the root cause in the sock-shop data exhibits distinct behaviors during normal and anomaly periods.
AutoMAP's performance is suboptimal in some cases due to the fact that it constructs the causal graph utilizing PC-algorithm, which, however, overlooks the temporal dependency.
Although we adhere to the settings used in RCD, $\Psi$-PC and RCD yield varying root causes in each run and perform worse on the sock-shop dataset.
CausalRCA neglects temporal information within time series data, leading to suboptimal performance in sock-shop.
The comparison of RUN and causal discovery-based approaches reveals the significance of temporal dependency.
RUN leverages positive pairs which capture the contextual information, and applies neural Granger causal discovery to consider the temporal dimension. 
This underscores the importance of considering temporal dependency, particularly in time series data within microservice systems.

As $\Psi$-PC and RCD cannot output the rank of the root cause, they are omitted for comparing performance in terms of MRR.
Table \ref{tab:MRR sock shop} illustrates the MRR of root cause analysis methods between RUN and $\epsilon$-diagnosis.
The table clearly illustrates that even though our performance in HR@1 for Memory leaks is the second best, it is still noticeable that the root cause maintains a relatively high ranking.
Furthermore, we observe that RUN outperforms $\epsilon$-diagnosis significantly, underscoring the impressive performance of RUN in identifying CPU hog incidents.

\begin{figure*}
    \centering
    \begin{minipage}[t]{\textwidth}
        \centering
        \begin{subfigure}[t]{0.33\textwidth}
            \tikz\node {\includegraphics[width=\textwidth]{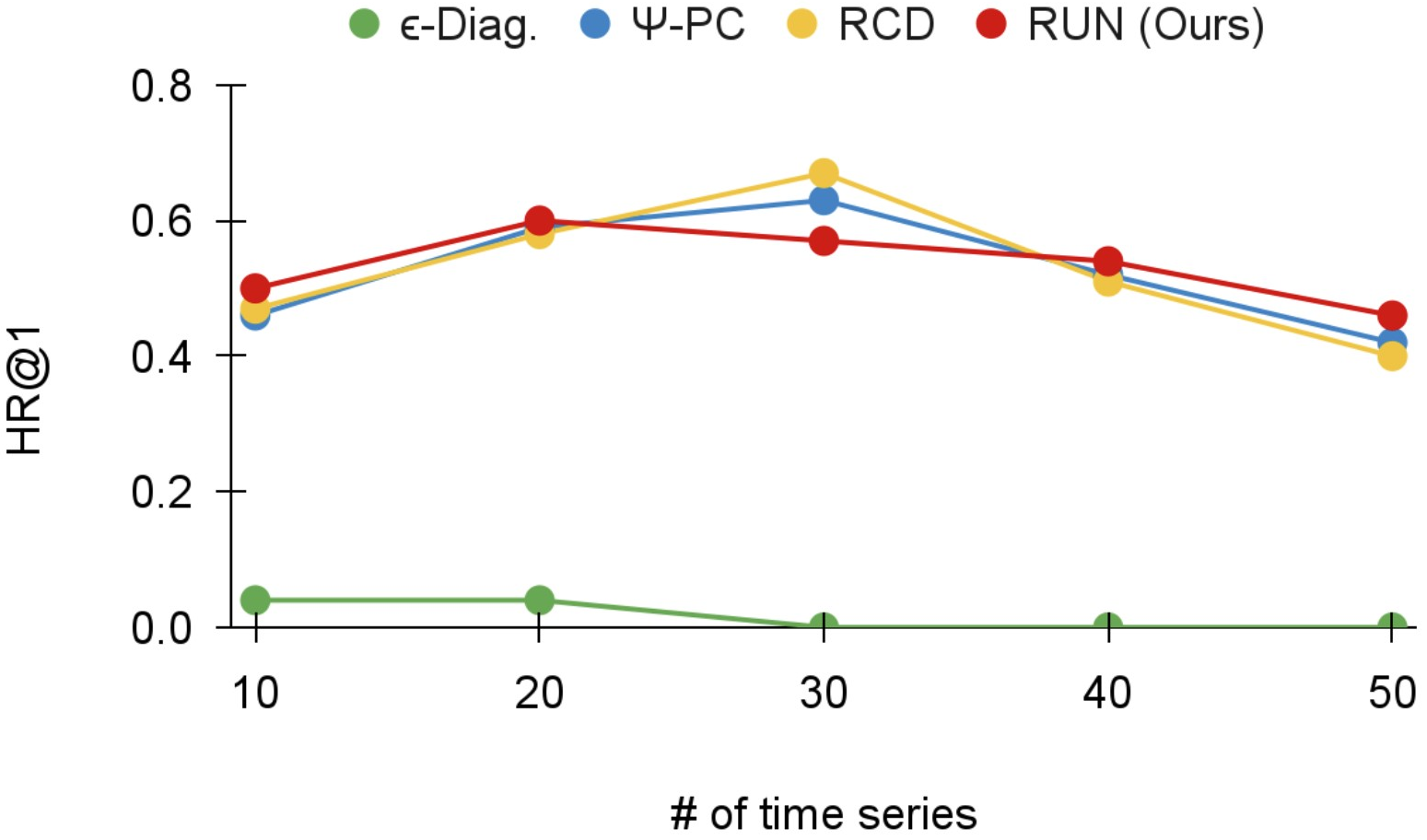}};
            \caption{HR@1}
            \label{fig:hr@1}
        \end{subfigure}
        \begin{subfigure}[t]{0.33\textwidth}
            \tikz\node{\includegraphics[width=\textwidth]{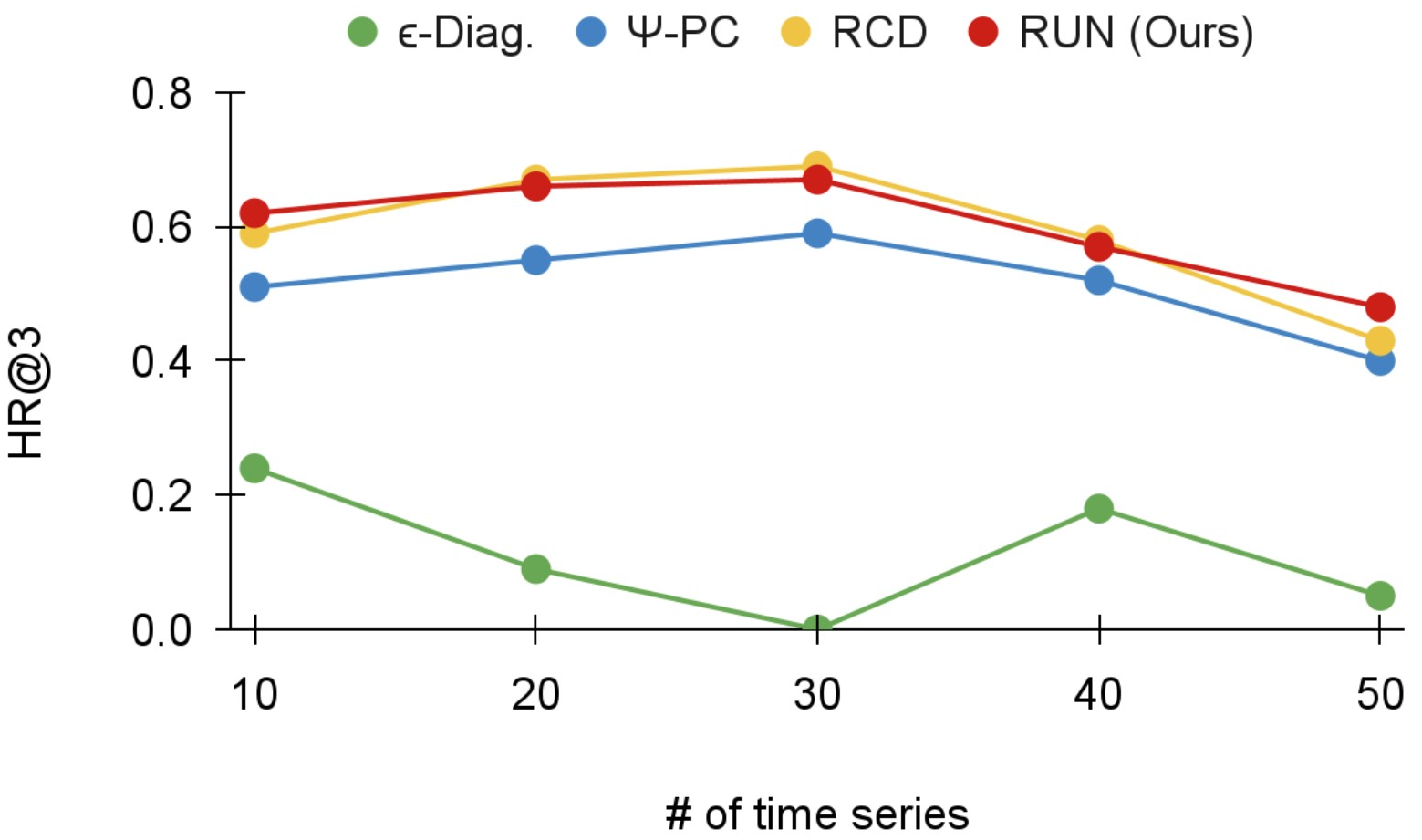}};
            \caption{HR@3}
            \label{fig:hr@2}
        \end{subfigure}
        \begin{subfigure}[t]{0.33\textwidth}
            \tikz\node{\includegraphics[width=\textwidth]{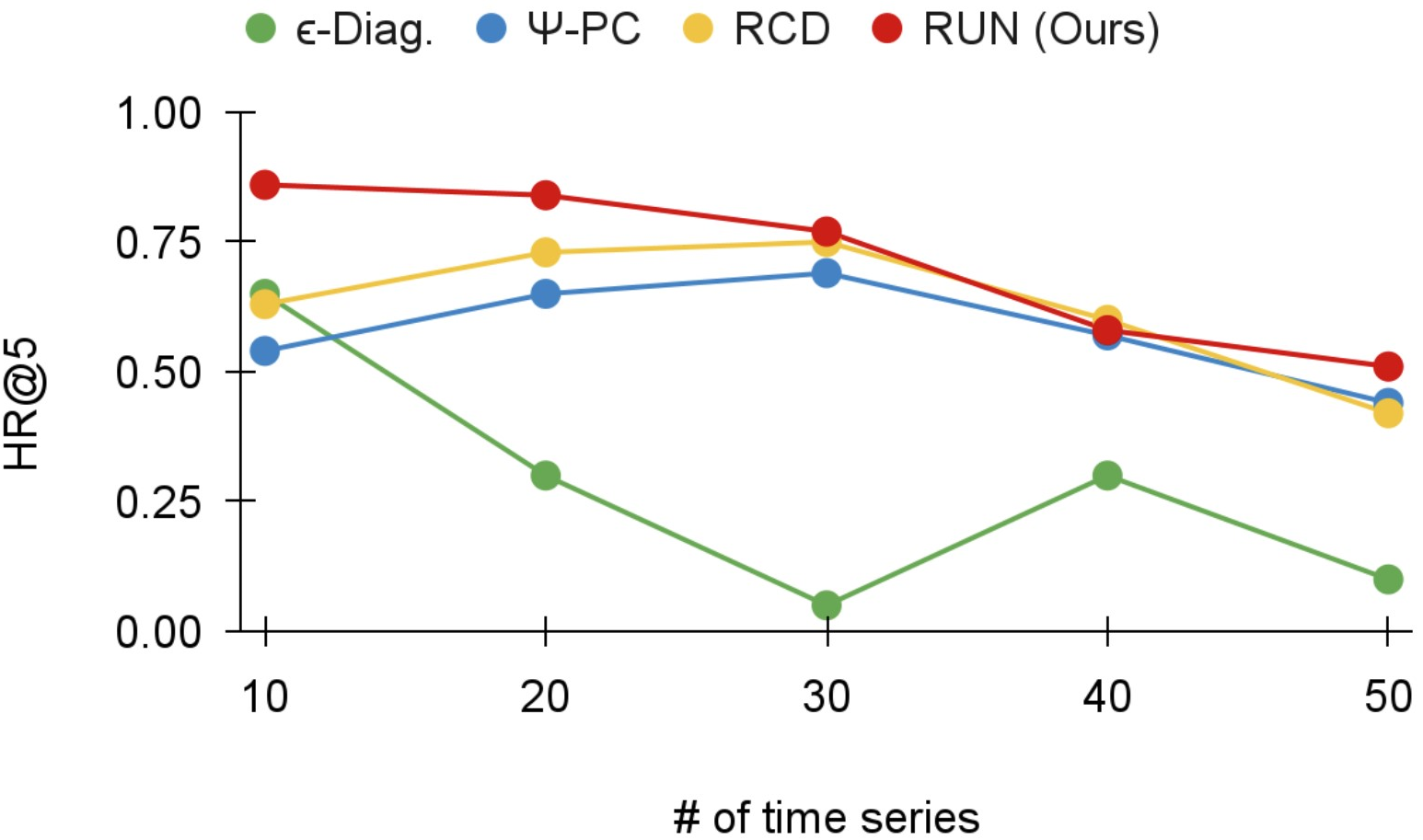}};
            \caption{HR@5}
            \label{fig:Hr@3}
        \end{subfigure}
    \end{minipage}
    \caption{HR@k on a synthetic dataset}
    \label{fig:synthetic}
\end{figure*}

\begin{figure}
    \includegraphics[width=1\linewidth, height=5cm]
  {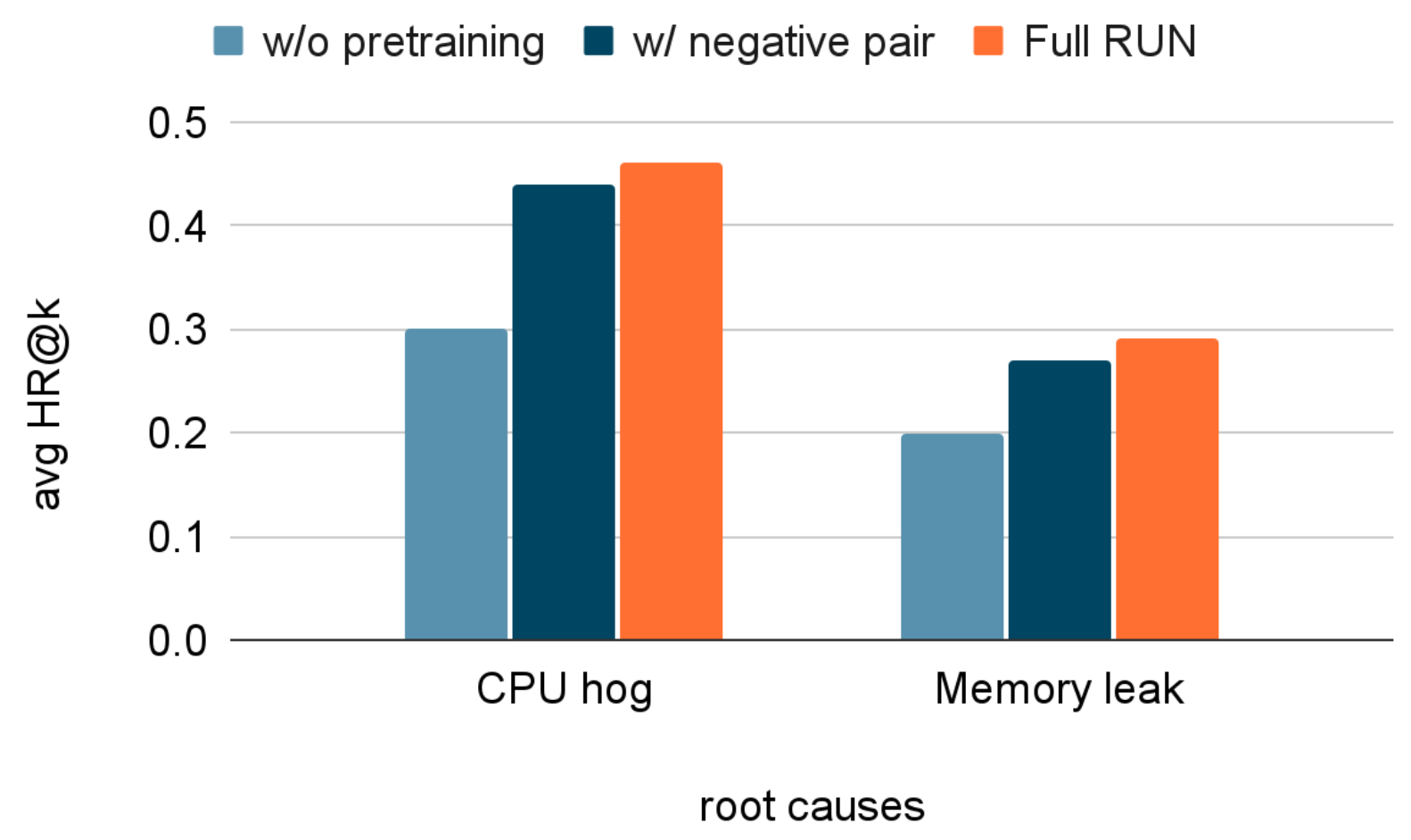}
    \caption{Ablation study removing the pre-training stage and incorporating the negative pairs in contrastive learning.}
    \label{fig:ablation study}
\end{figure}

\subsubsection{Synthetic Data}
Figure \ref{fig:synthetic} shows the HR@k of root cause analysis methods between RUN and baselines.
However, $\epsilon$-diagnosis falls short due to the fact that synthetic data lack the distinctive behaviors during both normal and anomaly periods that are characteristic of microservice-based data when anomalies occur.
Since the synthetic dataset is generated using the function of generateCPT in PyAgrum, it might not fully exhibit the characteristics of time series data.
Both $\Psi$-PC and RCD, on the other hand, learn causal graphs using the PC-algorithm from these synthetic datasets. 
As a result, they exhibit better performance on synthetic data when compared to the sock-shop data.
However, our model is specifically designed for performing root cause analysis within microservice systems consisting of abundant time series data, making it less suitable for application to the synthetic dataset. 
Nonetheless, our approach still demonstrates competitive performance with both RCD and $\Psi$-PC.

\subsection{Ablation Study}
To dissect the contributions arising from different designs of RUN, we conducted an ablation study by omitting the pre-training stage and incorporating the consideration of negative pairs in contrastive learning on sock-shop data for its microservice-based architecture. 
The results are presented in Figure \ref{fig:ablation study}.
As expected, removing the pre-training stage significantly results in a notable reduction in the performance of RUN.
This indicates that when performing root cause analysis via neural Granger causal discovery, incorporating contextual information can effectively elevate performance.
Nevertheless, excluding the incorporation of negative pairs in contrastive learning does not lead to a substantial decrease in performance.
We theorize that this finding can be ascribed to the limited temporal extent of the sock-shop data, potentially constraining the capacity to identify the multi-periodicity within the temporal domain.

\subsection{Case Study: Neural Granger Causal Discovery}
\begin{figure}
    \includegraphics[width=1\linewidth, height=3cm]
  {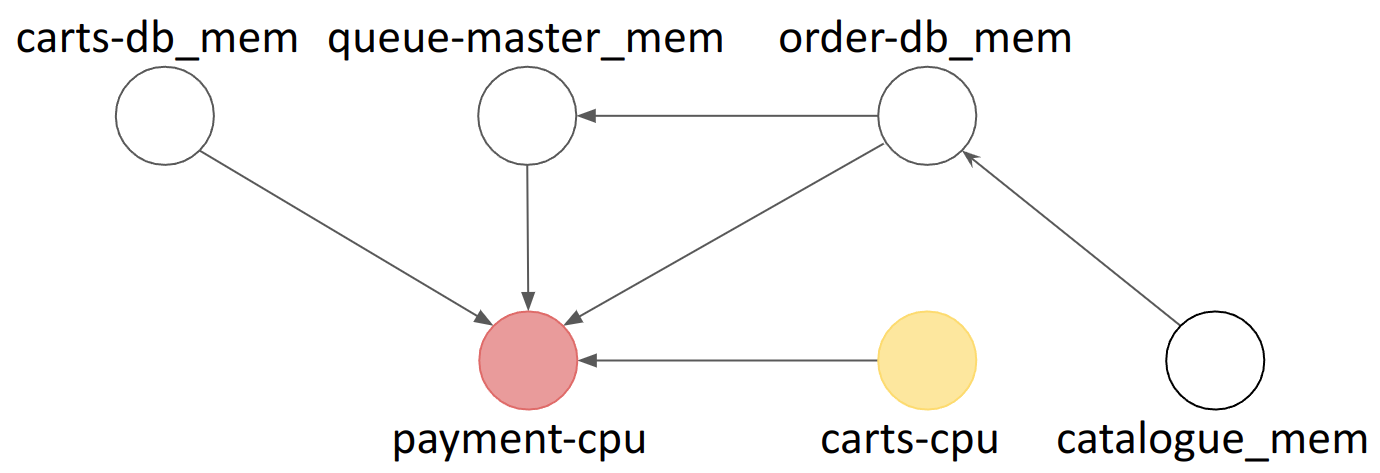}
    \caption{Construct the causal graph from the multivariate time series.}
    \label{fig:case study}
\end{figure}

In order to investigate the efficacy of neural Granger causal discovery, we sample a causal graph in which each node has a pathway to the root cause.
In Figure \ref{fig:case study}, the yellow node represents the trigger point and the red node represents the root cause.
Figure \ref{fig:case study} represents causal relationships that are associated with temporal order based on Granger causality.
To identify the root cause, each edge is reversed from the original causal graph for PageRank analysis.
According to the observation from \citet{DBLP:conf/kbse/WangWJHWKX21}, the root cause is a dangling node and impacts the trigger point.
Our approach validates the observation and successfully identifies the root cause from the causal graph.
It also illustrates the neural Granger causal discovery-based approach in explaining how to identify the root cause.

\section{Conclusion}
In this paper, we propose RUN for identifying the root causes in microservices.
Existing works apply PC-algorithm on time series data which leads to overlooking the temporal dimension of data. 
Distinct from these works, our proposed method is able to capture the temporal dependency.
Based on neural Granger causal discovery architecture, our model learns the graph structure using a time series forecasting model.
However, previous works on neural Granger causal discovery did not leverage the inherent contextual information in time series data.
Hence, contrastive learning without negative pairs is proposed to leverage the context of time series to enhance representations of time series.
The quantitative analysis carried out on both synthetic and sock-shop datasets showcases the effectiveness of our proposed approach compared to state-of-the-art baselines.
For our future work, we intend to enhance the scalability of our model to accommodate datasets of varying sizes. 
This improvement is crucial as it will allow us to not only build upon the experiments conducted on the sock-shop dataset but also extend our research to larger-scale and more representative datasets. These larger datasets will provide a more comprehensive understanding of our model's performance in real-world scenarios, further enriching the depth and applicability of our findings.

\bibliography{aaai24.bib}

\end{document}